\documentclass[10pt, conference, letterpaper]{IEEEtran}
\IEEEoverridecommandlockouts
\usepackage{cite}
\usepackage{amsmath,amssymb,amsfonts}
\usepackage{algorithmic}
\usepackage{graphicx}
\usepackage{textcomp}
\usepackage{xcolor}
\def\BibTeX{{\rm B\kern-.05em{\sc i\kern-.025em b}\kern-.08em
    T\kern-.1667em\lower.7ex\hbox{E}\kern-.125emX}}

\usepackage{mathtools}
\usepackage{booktabs,array,siunitx}
\usepackage[colorlinks=true, linkcolor=black, citecolor=black, urlcolor=black]{hyperref}

\begin{document}

\title{Learned Digital Over-the-Air Computing \\ for Federated Edge Learning
\thanks{This work was funded by UK Research and Innovation (UKRI) under the UK government’s Horizon Europe funding guarantee for projects DeepMARA (grant no. 101103430), 6G-GOALS (grant no. 101139232) and AI-R (European Research Council (ERC) Consolidator Grant, EP/X030806/1).}
}

\author{
\IEEEauthorblockN{Antonio Tarizzo, Mohammad Kazemi, and Deniz Gündüz}
\IEEEauthorblockA{Department of Electrical \& Electronic Engineering, Imperial College London, London, UK\\
\{antonio.tarizzo24, mohammad.kazemi, d.gunduz\}@imperial.ac.uk}
}

\maketitle
\begin{abstract}
Over-the-air (OTA) aggregation enables federated edge learning (FEEL) by exploiting the superposition property of the wireless channel to merge communication with computation, eliminating the need to schedule and decode devices individually. Analog OTA schemes transmit uncoded updates but are sensitive to noise, fading, and power misalignment, motivating more robust digital alternatives. However, state-of-the-art (SoTA) digital OTA designs that combine unsourced random access (URA) with compressed sensing struggle in the low signal-to-noise ratio (SNR) regimes common in Internet of Things (IoT) deployments, where symbol recovery and active-device estimation become unreliable. We propose a learned digital OTA framework that jointly trains a URA codebook with an unrolled approximate message passing (AMP)-based decoder. The learned decoder incorporates per-layer damping, residual scaling, temperature-controlled Bayesian denoising, and a lightweight convolutional neural network (CNN) refinement, while the codebook is optimised end-to-end through a factorised parameterisation. At near-perfect-aggregation accuracy, the proposed design extends the viable SNR range by approximately 7\,dB over the SoTA baseline at the same uplink overhead, and generalises across models, activity levels, and heterogeneous data.
\end{abstract}


\begin{figure*}[t]
    \centering
    \includegraphics[width=\textwidth]{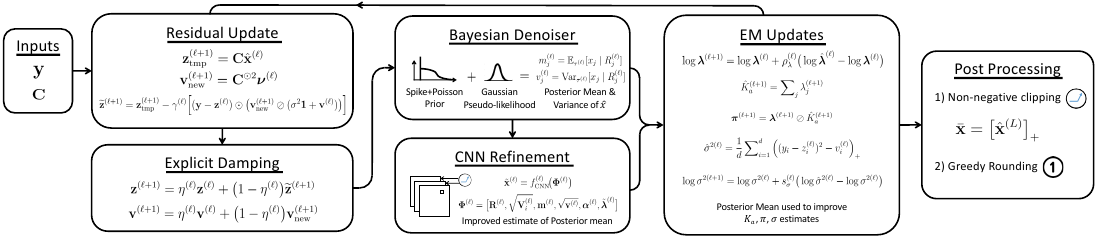}
    \vspace{-16pt}
    \caption{AMP-DA-Net decoder structure.}
    \vspace{-12pt}
    \label{fig:ampdanet}
\end{figure*}

\vspace{-3pt}
\section{Introduction}
\label{sec:intro}
Federated learning (FL) enables multiple clients to collaboratively train a shared model while keeping their data local \cite{mcmahan2017communication, konecny2016federated_strategies}. When deployed at the wireless edge across smartphones, Internet of Things (IoT) sensors, and autonomous systems, it must contend with the severe uplink bottleneck created by repeated transmission of model updates \cite{Chen:JSAC:21, jia2025comprehensive_survey}. Sparsification and quantisation reduce update size \cite{aji2017sparse,alistarh2017qsgd}, but OTA aggregation is more fundamental: simultaneous transmission and superposition merge communication with computation, cutting scheduling overhead and latency \cite{Amiri:TSP:20,Amiri:TWC:20}.

Analog OTA methods transmit uncoded updates directly but are sensitive to noise, fading, and power misalignment, limiting their practicality \cite{jia2025comprehensive_survey, sun2022_dynamic_scheduling_ota_fe_edge_learning}. Digital OTA provides an alternative by encoding updates before transmission, combining concurrent multiple access with the robustness of digital communication \cite{zhu2021onebit_over_the_air_aggregation}. However, existing digital OTA designs, such as massive digital over-the-air computation (MD-AirComp) proposed in \cite{qiao2024_massive_digital_ota_computation}, rely on hand-crafted compressed sensing decoders that struggle to converge in low signal-to-noise ratio (SNR) regimes common in IoT deployments, where symbol recovery and active-device estimation become unreliable. This motivates the development of more robust digital OTA frameworks that leverage learning-based codebook design and decoding to improve recovery and convergence without increasing uplink overhead.

\textbf{Related Work:} Early federated edge learning (FEEL) systems used reliable orthogonal access such as TDMA and OFDMA, but assigning distinct resources and decoding each device separately incurs scheduling and latency overhead that scales poorly. OFDMA-F$^2$L jointly optimises client selection, sub-channel allocation, and modulation, but retains this per-device cost \cite{hu2024_ofdmaF2L}.
Digital OTA instead exploits simultaneous transmission. One-bit digital OTA aggregation (OBDA) showed that FL can tolerate extreme quantisation, but its one-bit updates impose an accuracy ceiling \cite{zhu2021onebit_over_the_air_aggregation}. Frequency-shift keying with majority vote (FSK-MV) improves robustness to hardware impairments and non-coherent reception, yet has the same one-bit limitation \cite{sahin2021_distributed_learning_fsk_majority_vote}. MD-AirComp combines unsourced random access (URA), compressed sensing (CS), and vector quantisation (VQ) for higher-resolution aggregation without device identities \cite{polyanskiy2017_perspective_massive_random_access,URA_Survey,qiao2024_massive_digital_ota_computation}, but recovery degrades sharply at low SNR, preventing convergence when symbol and active-device estimates become unreliable. Type-based unsourced multiple access (TUMA) estimates both message support and multiplicities, aligning with our count-recovery objective \cite{ngo2024_tuma}.

\textbf{Contributions:} \textit{Low-SNR performance:} To our knowledge, the proposed AMP-DA-Net is the first learned digital OTA aggregation pipeline. It reaches near-perfect-aggregation accuracy at 3\,dB rather than approximately 10\,dB for MD-AirComp, without additional uplink overhead. \textit{Generalisation:} It remains stable across model architectures, activity levels, SNRs, and device data that are not independent and identically distributed (non-IID). \textit{Codebook design:} Jointly learning the URA codebook systematically outperforms fixed constructions, supporting end-to-end learned URA design.

\textbf{Notation:} Lower-case, bold lower-case, and bold upper-case letters denote scalars, vectors, and matrices, respectively. $\mathbf{I}_d$ denotes the $d\times d$ identity matrix. Superscripts in parentheses indicate the iteration or layer index. The operators $\odot$ and $\oslash$ denote element-wise multiplication and division, respectively. Expectation and variance are denoted by $\mathbb{E}[\cdot]$ and $\mathrm{Var}(\cdot)$.

\section{System Model}
\label{sec:system}
We consider a FEEL system with $K$ edge devices. In each round, a subset $\mathcal{S}_a$ of $K_a=|\mathcal{S}_a|$ active devices is sampled uniformly without replacement from the $K$ devices. Each device $k$ has a single antenna and local dataset $\mathcal{D}_k$. For simplicity, devices are assumed to have equal-sized datasets, $|\mathcal{D}_k| = B$, and equal computational capability. Although heterogeneity in data and hardware affects participation dynamics in practice, it does not alter the communication model and is abstracted away for clarity. A single-antenna base station (BS), co-located with the parameter server, performs uplink decoding, aggregation, and global-model broadcasting. Symbol-level synchronisation is assumed so that simultaneous transmissions align in time and frequency \cite{Sahin:PIMRC:25}. The distributed learning objective is
\vspace{-3pt}
\begin{equation}
  \min_{\mathbf{w}\in\mathbb{R}^W} f(\mathbf{w})
  \;=\;
  \frac{1}{K}\sum_{k=1}^{K} F_k(\mathbf{w}),
\end{equation}
where $\mathbf{w}\in\mathbb{R}^W$ is the global model and $F_k(\mathbf{w})$ is the empirical loss on $\mathcal{D}_k$. At round $t$, the BS broadcasts $\mathbf{w}^{t-1}$ to all devices. Each active device performs $E$ local stochastic gradient descent (SGD) steps to obtain $\mathbf{w}_k^t$, yielding
$\Delta \mathbf{w}_k^t = \mathbf{w}_k^t - \mathbf{w}^{t-1}$.
Each active device first forms the error-feedback-compensated update $\mathbf{s}_k^t$ and then transmits its compressed representation $\mathcal{Q}(\mathbf{s}_k^t)$, where
\begin{equation}
  \mathbf{s}_k^t = \Delta \mathbf{w}_k^t + \mathbf{e}_k^{t-1},
  \qquad
  \mathbf{e}_k^t = \mathbf{s}_k^t - \mathcal{Q}(\mathbf{s}_k^t).
\end{equation}
Here, $\mathcal{Q}(\cdot)$ denotes a generic compression operator. The BS aggregates the recovered compressed updates as
\begin{equation}
  \mathbf{s}^t
  \;=\;
  \frac{1}{K_a}\sum_{k\in\mathcal{S}_a}\mathcal{Q}(\mathbf{s}_k^t),
  \label{eq:aggregate}
\end{equation}
and updates the global model as
$\mathbf{w}^t = \mathbf{w}^{t-1} + \eta\,\mathbf{s}^t$.

\textbf{Sparse Recovery Formulation:}
In the uplink, each device divides its compressed update into fixed-length fragments, and each fragment is mapped to a codeword from a shared URA codebook $\mathbf{C}\in\mathbb{R}^{l\times n}$, whose columns correspond to the possible transmitted codewords. Here, $l$ is the transmitted codeword length and $n$ is the URA codebook size. In a given fragment slot, all active devices transmit their selected codewords simultaneously. This induces a sparse recovery problem. Let $\mathbf{x}\in\mathbb{N}_0^n$ denote the activity vector whose $i$-th entry counts how many active devices selected the $i$-th codeword. Then, $\mathbf{x}$ is sparse, non-negative, and integer-valued, with $\|\mathbf{x}\|_0 \ll n$ and $\sum_{i=1}^n x_i = K_a$. The received signal at the BS for a single fragment slot is
$\mathbf{y} = \mathbf{C}\mathbf{x} + \mathbf{n}$,
where $\mathbf{n}\sim\mathcal{N}(0,\sigma^2\mathbf{I}_l)$ is additive white Gaussian noise (AWGN). The BS must recover $\mathbf{x}$, or equivalently, the selected codeword counts, from these noisy linear measurements. This matches a compressed sensing problem with additional structure induced by the URA representation and device activity constraints, enabling simultaneous transmissions without preambles or explicit device identifiers.

\vspace{-3pt}
\section{Learned Digital AirComp Framework}
\label{sec:proposed}
The proposed framework integrates a learned encoder-decoder into the FEEL uplink. Trained offline and fixed for deployment, the communication layer yields a URA codebook and decoder reusable across rounds without retraining. In each round, devices perform local training and form the error-feedback-compensated update $\mathbf{s}_k^t$. This vector is split into $J$ length-$d$ fragments $\mathbf{s}_{k,j}^t\in\mathbb{R}^d$, $j=1,\ldots,J$. Each fragment is quantised by nearest-neighbour search over the vector quantisation codebook $\mathbf{Q}\in\mathbb{R}^{d\times n}$ broadcast by the BS,
\vspace{-3pt}
\begin{equation}
  \hat{\mathbf{q}}
  =
  \arg\min_{\mathbf{q}\in\mathbf{Q}}
  \|\mathbf{s}_{k,j}^t - \mathbf{q}\|_2.
  \label{eq:vq}
\end{equation}
The selected centroid index is mapped to its codeword in the shared URA codebook $\mathbf{C}$ and transmitted alongside those selected by other active devices. The BS receives the superposition and applies the learned decoder to estimate $\hat{\mathbf{x}}$. It then reconstructs the quantised fragments, forms the aggregate update in \eqref{eq:aggregate}, and updates the global model. The updated model and next-round quantisation codebook are broadcast to the devices.

\vspace{-3pt}
\subsection{Quantisation Codebook Construction}
\label{quant_codebook}
At the start of each round, the BS trains on its local dataset and fragments its update into length-$d$ vectors. Clustering these fragments using k-means$++$ \cite{arthur2007_kmeanspp} yields $n$ centroids forming $\mathbf{Q}\in\mathbb{R}^{d\times n}$. The BS then quantises its fragments and records the centroid assignment counts $N_i$, producing the normalised popularity distribution
\vspace{-3pt}
\begin{equation}
  \hat{\pi}_i = \frac{N_i}{\sum_i N_i}.
\end{equation}
The codebook is then ordered from most to least popular before being broadcast. The motivation for this ordering is that the BS popularity distribution provides a useful proxy for the codeword usage of other devices. Applying the same ordering across devices standardises the expected input distribution presented to the URA codebook across rounds and learning tasks. In addition, the decoder can exploit the fact that earlier codewords are more likely to be active, effectively using codeword popularity as prior information during recovery.

\subsection{AMP-DA-Net (Learned Decoder)} \label{amp_da_net}
AMP-DA-Net unrolls the AMP-based digital aggregation (AMP-DA) decoder used in MD-AirComp, drawing on AMP, generalised AMP (GAMP), and AMP-Net \cite{rangan2011_generalized_amp,zhang2021_ampnet,qiao2024_massive_digital_ota_computation}. As illustrated in Fig.~\ref{fig:ampdanet}, each fixed-depth layer alternates measurement-domain residual updates with codeword-domain denoising while learning residual scaling and damping, temperature-controlled Bayesian denoising, convolutional neural network (CNN) refinement, and expectation-maximisation (EM)-style prior and noise updates. Fragment slots are decoded independently. Hence, we describe one fragment $j\in\{1,\ldots,J\}$ and use $i\in\{1,\ldots,n\}$ to index codewords.

\begin{figure*}[t]
    \vspace{-12pt}
    \centering
    \begin{minipage}[t]{0.485\textwidth}
        \centering
        \includegraphics[width=\linewidth]{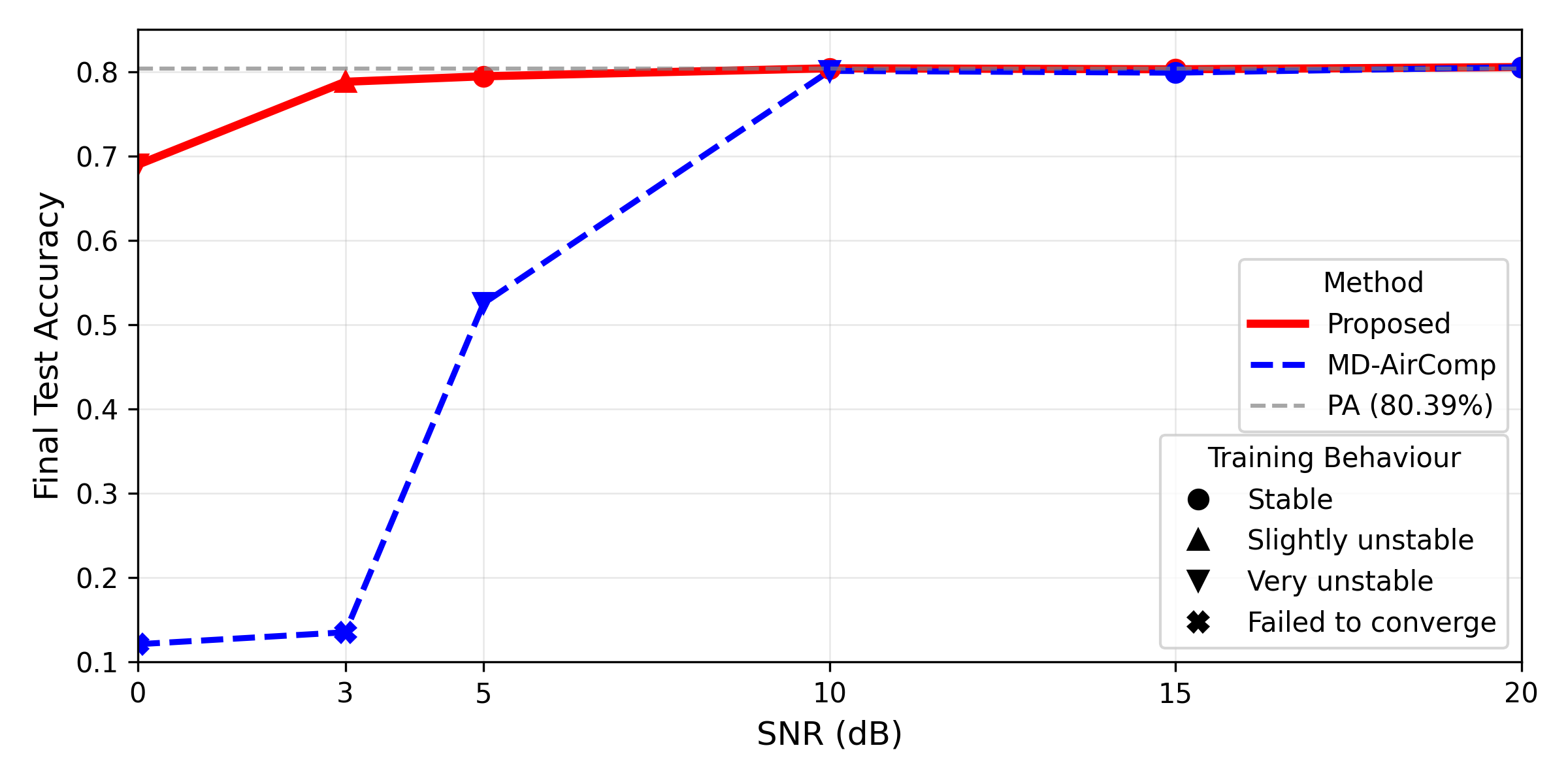}
        \vspace{-22pt}
        \caption{Final test accuracies and training stability.}
        \label{fig:fed_acc}
    \end{minipage}
    \hfill
    \begin{minipage}[t]{0.485\textwidth}
        \centering
        \includegraphics[width=0.95\linewidth]{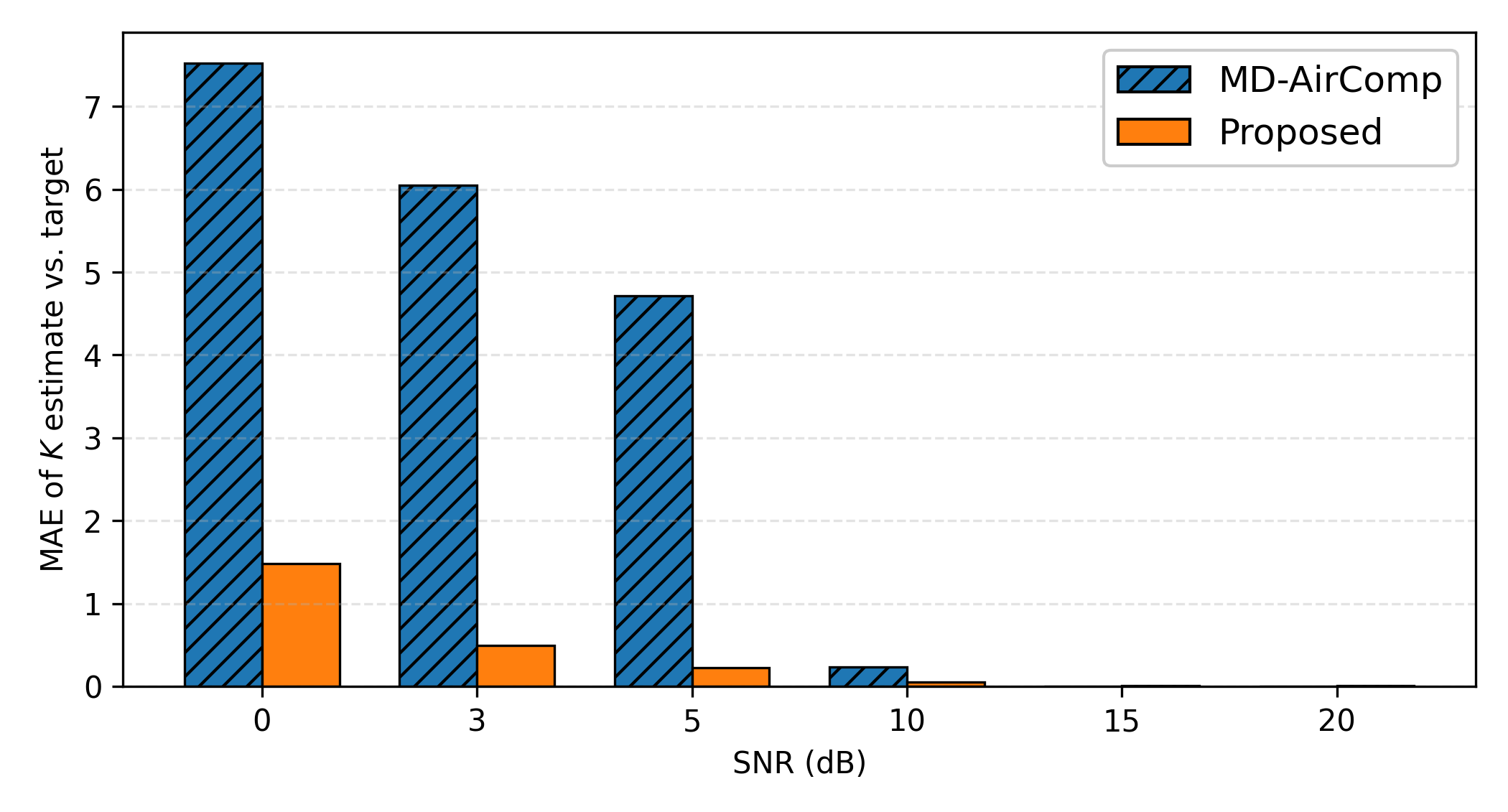}
        \vspace{-11pt}
        \caption{Mean absolute error for $\hat{K}_a$ (unrounded).}
        \label{fig:mae}
    \end{minipage}
    \vspace{-10pt}
\end{figure*}

\vspace{3pt}
\noindent
\textbf{Initialisation:}
For each fragment, the decoder runs for $L$ layers. At layer $\ell$, it maintains the codeword-domain estimate $\hat{\mathbf{x}}^{(\ell)}\in\mathbb{R}^{n}$ and variance proxy $\boldsymbol{\nu}^{(\ell)}\in\mathbb{R}^{n}$, together with the measurement-domain estimate $\mathbf{z}^{(\ell)}\in\mathbb{R}^{l}$ and variance proxy $\mathbf{v}^{(\ell)}\in\mathbb{R}^{l}$. The residual is
$\mathbf{r}^{(\ell)} = \mathbf{y} - \mathbf{z}^{(\ell)}$.
These variables are initialised as
\begin{equation}
\hat{\mathbf{x}}^{(0)}=\mathbf{0}, \qquad
\boldsymbol{\nu}^{(0)}=\mathbf{1}, \qquad
\mathbf{z}^{(0)}=\mathbf{y}, \qquad
\mathbf{v}^{(0)}=\mathbf{1}.
\end{equation}

\noindent
\textbf{Output Block:} 
The output block projects codeword-domain statistics through the codebook to update the measurement-domain mean and variance:
\begin{equation}
\mathbf{z}_{\mathrm{tmp}}^{(\ell+1)} = \mathbf{C}\hat{\mathbf{x}}^{(\ell)},
\qquad
\mathbf{v}_{\mathrm{new}}^{(\ell+1)} = \mathbf{C}^{\odot 2}\boldsymbol{\nu}^{(\ell)}.
\end{equation}
Here, $\mathbf{C}^{\odot 2}$ denotes the element-wise square of $\mathbf{C}$.
Defining the effective measurement variance as $\mathbf{d}^{(\ell)}=\sigma^2\mathbf{1}+\mathbf{v}^{(\ell)}$,
the output update is
\begin{align}
\widetilde{\mathbf{z}}^{(\ell+1)}
&=
\mathbf{z}_{\mathrm{tmp}}^{(\ell+1)}
-\gamma^{(\ell)}
\Big[
\mathbf{r}^{(\ell)} \odot
\big(
\mathbf{v}_{\mathrm{new}}^{(\ell+1)} \oslash \mathbf{d}^{(\ell)}
\big)
\Big],\\
\mathbf{z}^{(\ell+1)}
&=
\eta^{(\ell)}\mathbf{z}^{(\ell)}
+\big(1-\eta^{(\ell)}\big)\widetilde{\mathbf{z}}^{(\ell+1)},\\
\mathbf{v}^{(\ell+1)}
&=
\eta^{(\ell)}\mathbf{v}^{(\ell)}
+\big(1-\eta^{(\ell)}\big)\mathbf{v}_{\mathrm{new}}^{(\ell+1)},
\end{align}
where $\gamma^{(\ell)}$ is a learnable residual-scaling parameter, restricted to $[0.3,2]$\footnote{Allowing values slightly above $1$ introduces controlled instability, shown to improve expressiveness and convergence in learned iterative decoders \cite{nareddy2025_intriguing_learnt_matrices}.}, and $0<\eta^{(\ell)}<1$ is a learnable damping factor.

\vspace{3pt}
\noindent
\textbf{Input Block:} 
The input block performs codeword-domain denoising through a scalar pseudo-channel approximation. Define the per-measurement precision vector
\begin{equation}
\boldsymbol{\kappa}^{(\ell)} =
\beta^{(\ell)} \odot (\mathbf{d}^{(\ell)})^{-1},
\end{equation}
where $\beta^{(\ell)}$ is a learnable per-layer scaling factor. Projecting this through the codebook yields the pseudo-channel precision, variance, and matched-filter term 
\begin{align}
\boldsymbol{\psi}^{(\ell)} &= \mathbf{C}^{\odot2\,\top}\boldsymbol{\kappa}^{(\ell)}, 
& \mathbf{V}^{(\ell)} &= \big(\boldsymbol{\psi}^{(\ell)}\big)^{-1}.
\end{align}
\begin{equation} 
\boldsymbol{\rho}^{(\ell)} = \mathbf{C}^\top\big(\boldsymbol{\kappa}^{(\ell)}\odot \mathbf{r}^{(\ell)}\big). 
\end{equation}
The resulting pseudo-observation in vector form and its scalar equivalent are
\begin{equation}
\mathbf{R}^{(\ell)}
=
\hat{\mathbf{x}}^{(\ell)}
+
\boldsymbol{\rho}^{(\ell)} \oslash \boldsymbol{\psi}^{(\ell)},
\label{eq:ampdanet_pseudo_channel}
\end{equation}
\begin{equation}
R_i^{(\ell)} = x_i + w_i,
\qquad
w_i \sim \mathcal{N}\!\big(0,(\mathbf{V}^{(\ell)})_i\big).
\end{equation}
Each coefficient is denoised using the spike-and-slab prior
\begin{equation}
p(x_i) = (1-\alpha_i)\delta_0 + \alpha_i \,\mathrm{Pois}(\lambda_i),
\end{equation}
reflecting that most codewords are inactive and active ones have non-negative integer counts. Combining this prior with the scalar Gaussian pseudo-channel yields the tempered posterior moments
\begin{align}
m_i^{(\ell)} &= \mathbb{E}_{\tau^{(\ell)}}[x_i \mid R_i^{(\ell)}], 
& \sigma_i^{(\ell)} &= \mathrm{Var}_{\tau^{(\ell)}}[x_i \mid R_i^{(\ell)}],
\end{align}
where $\tau^{(\ell)}>0$ is a learnable per-layer temperature that scales the log-posterior as $\log p_{\tau^{(\ell)}}(x_i \mid R_i^{(\ell)}) \propto (\tau^{(\ell)})^{-1} \log p(x_i \mid R_i^{(\ell)})$, sharpening ($\tau < 1$) or smoothing ($\tau > 1$) the distribution. Collecting these into vectors $\mathbf{m}^{(\ell)}$ and $\mathbf{\sigma}^{(\ell)}$, the variance state is updated as $\boldsymbol{\nu}^{(\ell+1)} = \mathbf{\sigma}^{(\ell)}$. Then, to refine the Bayesian estimate, a small one-dimensional CNN operating across the $n$ codeword indices (with the 6 input quantities treated as channels) takes the feature map
\begin{equation}
\boldsymbol{\Phi}^{(\ell)}=
\big[
\mathbf{R}^{(\ell)},\,
\sqrt{\mathbf{V}^{(\ell)}},\,
\mathbf{m}^{(\ell)},\,
\sqrt{\mathbf{\sigma}^{(\ell)}},\,
\boldsymbol{\alpha}^{(\ell)},\,
\tilde{\boldsymbol{\lambda}}^{(\ell)}
\big],
\end{equation}
Here, $\tilde{\lambda}_i^{(\ell)} = (\log \lambda_i^{(\ell)} - \mu^{(\ell)}) / s^{(\ell)}$ is the normalised log-rate, with mean $\mu^{(\ell)}$ and standard deviation $s^{(\ell)}$ computed across codewords. A learnable gate $0\le \zeta^{(\ell)}\le 1$ blends the CNN output with the Bayesian posterior mean:
\begin{equation}
\tilde{\mathbf{x}}^{(\ell)} = f_{\mathrm{CNN}}^{(\ell)}\!\big(\boldsymbol{\Phi}^{(\ell)}\big)
\end{equation}
\begin{equation}
\hat{\mathbf{x}}^{(\ell+1)}
=
\big(1-\zeta^{(\ell)}\big)\mathbf{m}^{(\ell)}
+
\zeta^{(\ell)}\tilde{\mathbf{x}}^{(\ell)}.
\end{equation}

\noindent
\textbf{Expectation-Maximisation (EM) Updates:}
\phantomsection\label{em_updates}
Each layer uses damped EM-style updates to adapt the latent prior across rounds and operating conditions. The posterior mean proposes Poisson rates merged with the current log-domain estimate:
\begin{equation}
\log \boldsymbol{\lambda}^{(\ell+1)}
=
\log \boldsymbol{\lambda}^{(\ell)}
+
\rho_\lambda^{(\ell)}
\big(
\log \mathbf{m}^{(\ell)}
-
\log \boldsymbol{\lambda}^{(\ell)}
\big).
\end{equation}
Here, $0<\rho_\lambda^{(\ell)}<1$ is learned or confidence-controlled. The activity estimate and popularity distribution are given by
\begin{align}
\hat{K}_a^{(\ell+1)} &= \sum_i \lambda_i^{(\ell+1)}, 
& \boldsymbol{\pi}^{(\ell+1)} &= \frac{\boldsymbol{\lambda}^{(\ell+1)}}{\hat{K}_a^{(\ell+1)}},
\end{align}
and the Bernoulli spike mass is refreshed via
\begin{equation}
\alpha_i^{(\ell+1)} \approx 1-e^{-\lambda_i^{(\ell+1)}}.
\end{equation}
The noise variance $\sigma^2$ is also refined with an EM-inspired residual-energy update in the log-domain, as in \cite{qiao2024_massive_digital_ota_computation}. In this way, the decoder is able to adapt to varying SNRs, activity levels, and popularity profiles.

\noindent
\textbf{Post-processing:}
\phantomsection\label{post_process}
A final correction step enforces the known structure of the activity vector. The decoder output is first clipped to be non-negative, and then projected onto the set of non-negative integer vectors whose entries sum to the estimated activity level $\hat{K}_a$. This is implemented using a greedy rounding procedure that approximately minimises $\|\hat{\mathbf{x}}-\mathbf{x}\|_2$ subject to $\mathbf{x}\ge 0$ and $\sum_i x_i=\hat{K}_a$. The resulting count vector is then de-quantised to recover the transmitted fragments, which are aggregated to produce the global update.

\noindent
\textbf{Dataset Collection:}
\phantomsection\label{sec:dataset_collection}
Training data are generated using a perfect-aggregation FEEL pipeline with error feedback, popularity ordering, and varying activity levels. Device codeword indices are collected as target count vectors $\mathbf{x}$ using the FEEL and heterogeneity setting of Section~\ref{sec:results}, thereby capturing non-uniform codeword usage, per-round activity variation, and device-level statistical bias.

\noindent
\textbf{Codebook Representation:}
\phantomsection\label{sec:codebook_rep}
The URA codebook is jointly learned with the decoder using a two-matrix parameterisation $\mathbf{C}_{\mathrm{syn}}=\mathbf{W}^{\top}\mathbf{D}$,
where $\mathbf{D}\in\mathbb{R}^{l\times n}$ stores base codewords and $\mathbf{W}\in\mathbb{R}^{l\times l}$ is a learned mixing transform. Following the SimVQ parameterisation \cite{SimVQ}, this improves gradient flow and stabilises training compared to learning $\mathbf{C}$ directly. Columns of $\mathbf{C}_{\mathrm{syn}}$ are renormalised to unit $\ell_2$ norm after each update to maintain equal transmit power. Gaussian, Bernoulli, and data-driven initialisations were tested for $\mathbf{D}$, with the best choice selected empirically, while $\mathbf{W}$ was initialised as identity.

\noindent
\textbf{Loss Function:}
\phantomsection\label{sec:loss_func}
The training objective combines four terms: (i) an activity-reconstruction mean-squared error (MSE) $\|\hat{\mathbf{x}}-\mathbf{x}\|_2^2$, (ii) a sparsity-promoting $\ell_1$ penalty normalised by the ground-truth scale, (iii) an orthogonality regulariser $\|\mathbf{W}^{\top}\mathbf{W}-\mathbf{I}_l\|_F^2$ to keep the learned transform well-conditioned, and (iv) an active-device estimation loss $(\hat{K}_{a}-K_{a})^2$. These terms are balanced by hyperparameters $\lambda_1$, $\lambda_W$, and $\lambda_K$.

\noindent
\textbf{Hyper-parameters:}
\phantomsection\label{sec:hyperparameter}
Pre-training used $256{,}000$ samples for training, $64{,}000$ for validation, and $20{,}000$ held-out samples for testing. The batch size was $64$, with a maximum of $500$ epochs and early stopping with patience $20$. Optimisation used Adam with learning rate $10^{-4}$, halved when validation loss plateaued for $10$ epochs. The decoder used $10$ unrolled layers, each with a one-dimensional CNN denoiser with $32$ filters and kernel size $3$. The loss weights were set to $\lambda_1=0.01$, $\lambda_W=0.001$, and $\lambda_K=0.01$.

\noindent
\textbf{Computational Complexity:}
AMP-DA-Net is linear in layer count and codebook size, dominated by parallel matrix-vector operations. CNN refinement is linear in codebook size and kernel width and approximately quadratic in filter count. Offline training and fixed-depth inference give consistent latency.

\section{Simulation Results}
\label{sec:results}
The proposed method is compared against the MD-AirComp baseline\footnote{All code is available at \url{https://github.com/tonytarizzo/AMP-DA-Net}}. Each device holds a local subset of CIFAR-10, split into $20\%$ IID and $80\%$ non-IID data. Specifically, $10{,}000$ samples were assigned randomly across devices, while the remaining $40{,}000$ were label-sorted into contiguous shards and distributed sequentially to induce heterogeneity. In each round, the number of active devices was drawn uniformly from $K_a\in\{7,\ldots,13\}$, after which $\mathcal{S}_a$ was sampled uniformly without replacement from the $K=40$ devices. The global model was updated using Federated Averaging (FedAvg), with a fragment dimension $d=20$, URA codeword length $l=64$, and codebook size $n=128$.

Figure~\ref{fig:fed_acc} shows that the gain is most pronounced at low SNR, where AMP-DA-Net maintains stable convergence while MD-AirComp fails to recover sufficiently accurate updates. AMP-DA-Net reaches the near-perfect-aggregation operating point at 3\,dB, whereas MD-AirComp requires approximately 10\,dB, representing a 7\,dB extension without additional uplink overhead. At moderate and high SNR, it matches MD-AirComp and remains close to the perfect-aggregation benchmark.

The accuracy of active-device estimation is also substantially improved. As shown in Fig.~\ref{fig:mae}, the mean absolute error of $\hat{K}_a$ remains below approximately $0.5$ over a much wider SNR range than MD-AirComp, meeting the absolute-error tolerance needed for stable scaling in \eqref{eq:aggregate}. This is important because underestimating $K_a$ amplifies the global update and can destabilise training, whereas overestimating it attenuates the update and slows convergence.

\begin{figure}[t]
    \vspace{-10pt}
    \centering
    \includegraphics[width=0.79\columnwidth]{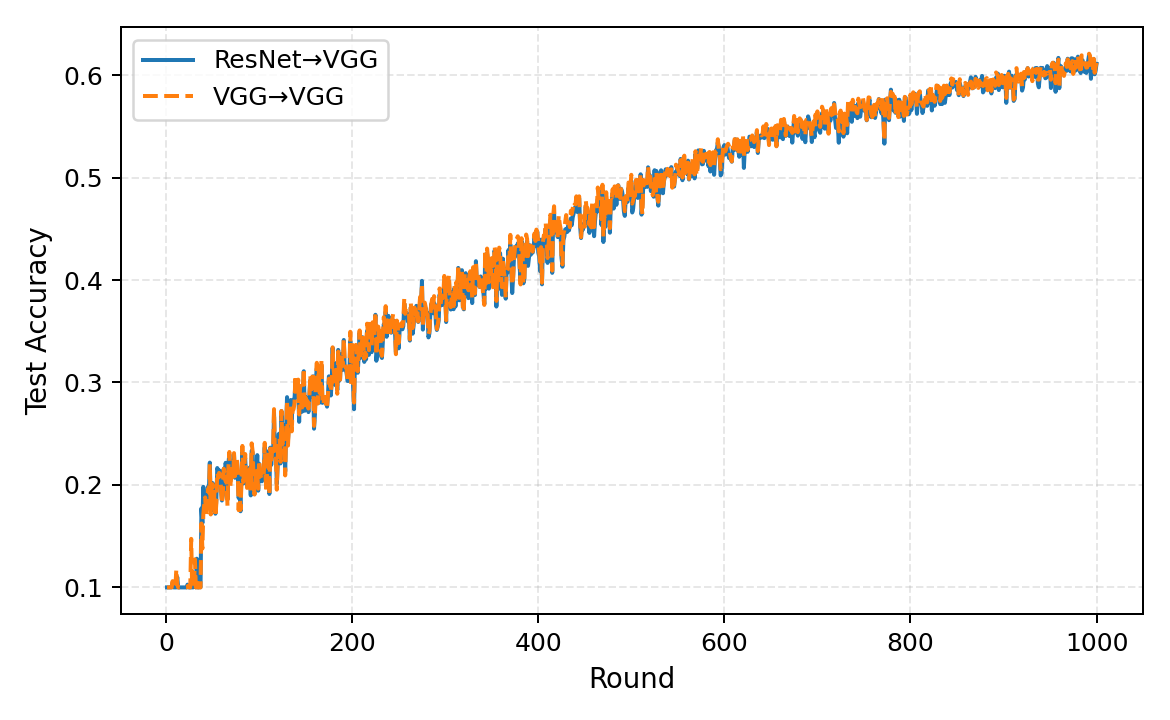}
    \vspace{-12pt}
    \caption{VGG vs ResNet-trained communication model comparison.}
    \label{fig:vgg_test}
    \vspace{-11pt}
\end{figure}

\phantomsection\label{sec:generalisation}
To evaluate generalisation, we trained two instances of the communication layer using datasets generated from different global model architectures: a ResNet-20 ($269{,}722$ parameters) and a compact VGG-style CNN with six convolutional layers ($288{,}298$ parameters). Both were deployed with the VGG architecture at $10$ dB SNR, so their accuracies are directly comparable to each other rather than to the ResNet-based results in Figs.~\ref{fig:fed_acc} and \ref{fig:mae}. As shown in Fig.~\ref{fig:vgg_test}, their convergence is essentially indistinguishable, indicating that communication training data generated with ResNet transfer to VGG deployment. This robustness is a structural consequence of the pipeline design: quantisation and URA encoding reduce the uplink task to recovering discrete codeword counts, decoupling the communication layer from the specifics of the global model, while popularity ordering stabilises the codeword usage across tasks and rounds.

\begin{table}[t]
\vspace{-3pt}
\centering
\caption{Decoder-level sparse recovery accuracies at $5\,\mathrm{dB}$.}
\label{tab:codebook_acc}
\vspace{-3pt}
\renewcommand{\arraystretch}{0.95}
\setlength{\tabcolsep}{4pt}
\begin{tabular}{lccc}
\toprule
Codebook & Initialisation & Ordering & Accuracy \\
\midrule
Learned & Data-driven & Popularity & \textbf{0.949} \\
Learned & Gaussian    & Popularity & 0.932 \\
Learned & Data-driven & None       & 0.877 \\
Fixed   & Data-driven & None       & 0.666 \\
Fixed   & Gaussian    & None       & 0.696 \\
Fixed   & Bernoulli   & None       & 0.689 \\
\bottomrule
\end{tabular}
\vspace{-13pt}
\end{table}

Table~\ref{tab:codebook_acc} isolates the effects of codebook learning, initialisation, and ordering at $5\,\mathrm{dB}$. Learned codebooks are jointly optimised, while fixed codebooks retain their initial construction. Initialisation is data-driven from the empirical training distribution or random Gaussian/Bernoulli. Popularity ordering uses the BS estimate $\hat{\boldsymbol{\pi}}$, while ``None'' retains the original order. Accuracy is the clipped normalised-$\ell_1$ score $[1-\|\hat{\mathbf{x}}-\mathbf{x}\|_1/K_a]_+$, averaged over held-out fragments. Without ordering, learning improves accuracy from $0.696$ for the best fixed codebook to $0.877$. Popularity ordering lifts this to $0.949$, versus $0.932$ with Gaussian initialisation, supporting end-to-end learning over a fixed random codebook.

\section{Conclusions}
\label{sec:future_work}
AMP-DA-Net jointly learns an unrolled AMP decoder and URA codebook for digital OTA aggregation. Under AWGN, it reaches the near-perfect-aggregation final-accuracy operating point at $3\,\mathrm{dB}$ rather than approximately $10\,\mathrm{dB}$ for MD-AirComp, extending low-SNR operation by $7\,\mathrm{dB}$ without additional uplink overhead. Experimental results show that the learned communication layer generalises across model architectures, activity levels, and highly non-IID data distributions, while learned codebooks consistently outperform fixed constructions. These findings support end-to-end learned digital OTA design beyond the FEEL setting considered here. The present framework assumes AWGN channels and single-antenna terminals. Extensions to block-fading and multi-antenna channels, together with formal state-evolution and convergence analysis, are immediate directions for future work.

\vfill
\null
\newpage
\bibliographystyle{IEEEtran}
\bibliography{main}

\end{document}